\pdfoutput=1
\documentclass[11pt,table]{article}

\usepackage[]{acl}

\usepackage{times}
\usepackage{latexsym}

\usepackage[T1]{fontenc}
\usepackage[utf8]{inputenc}

\usepackage{microtype}

\usepackage[english]{babel}
\usepackage{amsthm}
\theoremstyle{definition}
\newtheorem{definition}{Definition}[section]
\theoremstyle{remark}
\newtheorem{corpus}{Corpus Example}[section]
\theoremstyle{remark}

\usepackage{tabularx, colortbl, xcolor}
\definecolor{mygray}{gray}{0.95}
\definecolor{mycyan}{HTML}{005397}
\definecolor{myred}{HTML}{E13333}
\usepackage{soul}
\usepackage{easyReview}

\newcommand{\RomanNumeralCaps}[1]
    {\MakeUppercase{\romannumeral #1}}

\usepackage{tikz}
\usepackage{tikz-dependency}

\usepackage{subcaption}
\usepackage{amssymb}

\definecolor{p13}{HTML}{BFB5D7}
\definecolor{b14}{HTML}{BEA1A5}
\definecolor{y15}{HTML}{F0Cf61}
\newcolumntype{a}{>{\columncolor{p13}}l}
\usepackage{algorithm}
\usepackage{algpseudocode}

\usepackage{listings}

\title{Revision for Concision: A Constrained Paraphrase Generation Task}

\author{\textbf{Wenchuan Mu}\quad \textbf{Kwan Hui Lim}\\
Singapore University of Technology and Design \\
\texttt{\{wenchuan\_mu,kwanhui\_lim\}@sutd.edu.sg}\\
}

\begin{document}
\maketitle
\begin{abstract}
Academic writing should be concise as concise sentences better keep the readers' attention and convey meaning clearly. Writing concisely is challenging, for writers often struggle to revise their drafts. We introduce and formulate revising for concision as a natural language processing task at the sentence level. Revising for concision requires algorithms to use only necessary words to rewrite a sentence while preserving its meaning. The revised sentence should be evaluated according to its word choice, sentence structure, and organization. The revised sentence also needs to fulfil semantic retention and syntactic soundness. To aide these efforts, we curate and make available a benchmark parallel dataset that can depict revising for concision. The dataset contains 536 pairs of sentences before and after revising, and all pairs are collected from college writing centres. We also present and evaluate the approaches to this problem, which may assist researchers in this area.
\end{abstract}

\section{Introduction}
Concision and clarity\footnote{We treat \textit{concision} and \textit{conciseness} as equivalent, and \textit{clarity} as part of \textit{concision}} are important in academic writing as wordy sentences will obscure good ideas (Figure~\ref{fig:obscure}). Concise writing encourages writers to choose words deliberately and precisely, construct sentences carefully to eliminate deadword, and use grammar properly~\cite{stanford2021}, which often requires experience and time. A first draft often contains far more words than necessary, and achieving concise writing requires revisions~\cite{monash2020}. As far as we know, currently this revision process can only be done manually, or semi-manually with the help of some rule-based wordiness detectors~\cite{long_2013}. We therefore introduce and formulate revising for concision as a natural language processing (NLP) task and address it. In this study, we make the following contributions:

\begin{figure}[t!]
    \centering
    \includegraphics[width=0.46\textwidth]{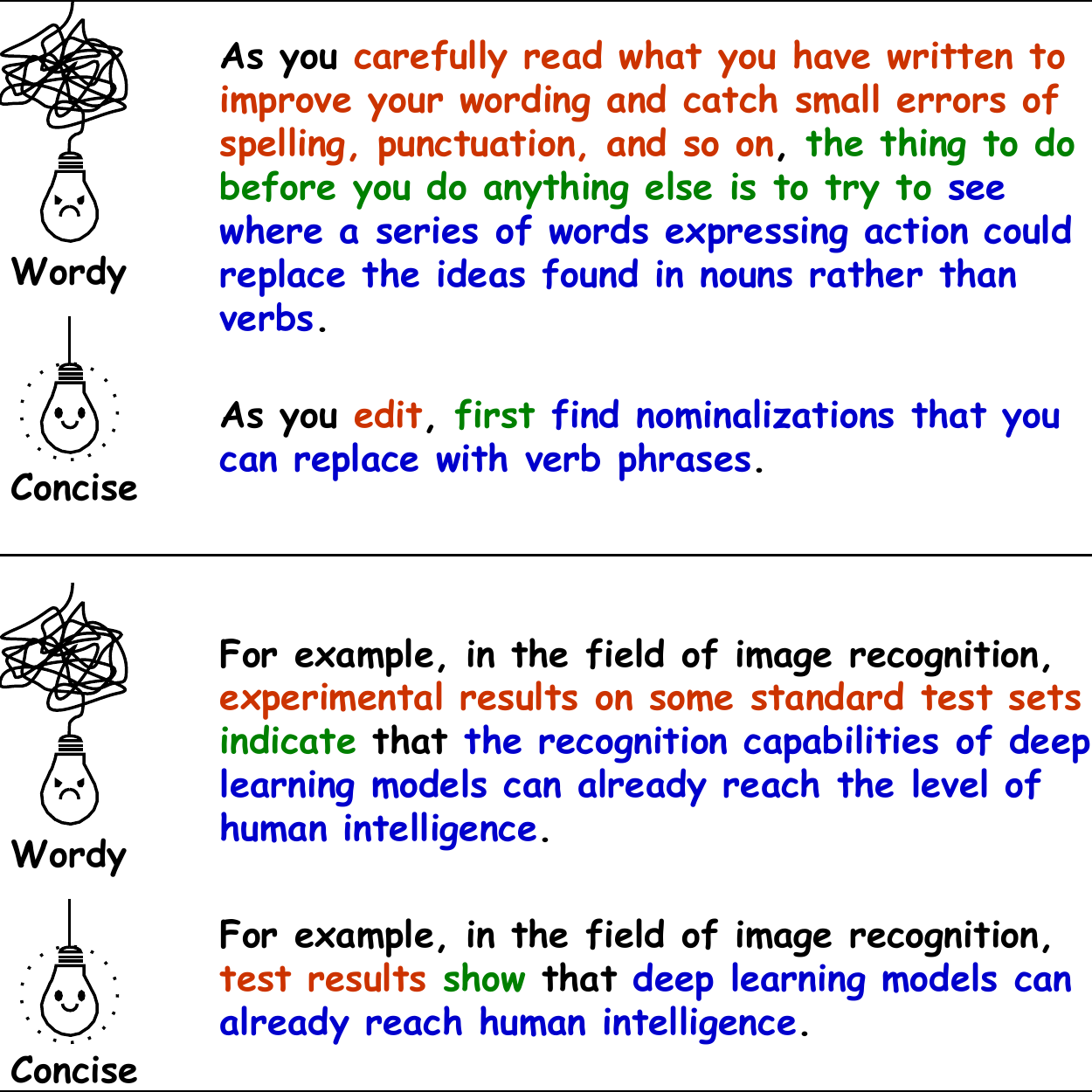}
    \caption{Wordy sentences are more boring to read than concise sentences. But how do we turn lengthy sentences into concise ones? We show two examples. The above sentence pair is taken from the Purdue Writing Lab, which suggests how college students should succinctly revise their writing~\cite{purdue_writing_lab2021}. In the other example, the wordy sentence comes from a scientific paper~\cite{chen2020survey}, and its concise counterpart is predicted from the concise revisioner we developed (Section~\ref{sec:methods}). In each pair, text with the same colour delivers the same information.}
    \label{fig:obscure}
\end{figure}

\begin{enumerate}
    \item We formulate the revising for concision NLP task at the sentence level, which reflects the revising task in academic writing. We also survey the differences between this task and sentence compression, paraphrasing, etc.
    \item We release a corpus of 536 sentence pairs, curated from 72 writing centres and additionally coded with the various linguistic rules for concise sentence revision.
    \item We propose an gloss-based Seq2Seq approach to this problem, and conduct automatic and human evaluations. We observed promising preliminary results and we believe that our findings will be useful for researchers working in this area.
\end{enumerate}

\section{Problem Statement}
\label{sec:problem}
\subsection{Revision as an English Writing Task}

Concise writing itself is a lesson that is often emphasized in colleges, and revision is crucial in writing. The following definitions are helpful when we set out to formulate the task.

\begin{definition}[Concise]
Marked by brevity of expression or statement: free from all elaboration and superfluous detail~\cite{merriam_webster_concise}.
\end{definition}

\begin{definition}[Concise writing, English]
Writing that is clear and does not include unnecessary or vague/unclear words or language~\cite{auckland2021}.
\end{definition}

Revising for concision at paragraph level, or even article level, may be the best practice. However, sentence-level revising usually suffices. We focus on revising for concision at the sentence level now. Indeed, in many college academic writing tutorials, revisions for concision are for individual sentences, and this process is defined as follows.

\begin{definition}[Revise for concision at the sentence level, English\footnote{Adapted from notes of~\citet{purdue_writing_lab2021} Writing Lab and~\citet{rambo2019}}]\label{def:eng}
Study a sentence in draft, use specific strategies\footnote{Presented in Appendix (Table~\ref{tab:strategies}) as a periphery of this study.}  to edit the sentence concisely without losing meaning.
\end{definition}

If someone, such as a college student, wants to concisely modify a sentence, specific strategies (\emph{e.g.}, \emph{delete} weak modifiers, \emph{replace} phrasal verbs with single verbs, or \emph{rewrite} in active voice, etc.) tell us how to locate wordiness and how to edit it~\cite{purdue_writing_lab2021,waldenu2021,ualr2021,massey2021,monash2020}. The rule is to repeatedly detect wordiness and revise it until no wordiness is detected or it cannot be removed without adding new wordiness. The final product serves as a concise version of the original sentence, if it does not lose its meaning.

\subsection{Task Definition in NLP}
Now that we know how humans can revise a sentence, what about programs? Each strategy is clear to a trained college student, but not clear enough to program in code. On the one hand, existing verbosity detectors may suggest which part of a sentence is too "dense"~\cite{long_2013}, but fail to expose fine-grained wordiness details. On the other hand, how programs can edit sentences without losing their meaning remains challenging. In short, no existing program can generate well-modified sentences in terms of concision.

Eager for a program that revises sentences nicely and concisely, we set out to formulate this modification process as a sequence-to-sequence (Seq2Seq) NLP task. In this task, the input is any English sentence and the output should be its concise version. We define it as follows.

\begin{definition}[Revise for concision at the sentence level, NLP]\label{def:nlp}
Produce a sentence where minimum wordiness can be identified. (And,) the produced sentence delivers the same information as input does. (And,) the produced sentence is syntactically correct.

\end{definition}

As many other NLP tasks, \emph{e.g.}, machine translation, named-entity recognition, etc., Definition\,\ref{def:nlp} describes the product (text) of a process, not the process itself, \emph{i.e.}, how the text is produced. This perspective is different from that of Definition\,\ref{def:eng}.

Among the three components in Definition\,\ref{def:nlp}, both the first and the third are clear and self-contained. They are related to syntax; hence, at least human experts would think it straightforward to determine the soundness of a sentence on both.  For example, the syntax correctness of an English sentence will not be judged differently by different experts, unless the syntax itself changes. Unfortunately, the second component is neither clear nor self-contained. This component asks for information retention, which is a rule inherited from Definition\,\ref{def:eng}. Determining the semantic similarity between texts has long been challenging, even for human experts~\cite{rus-etal-2014-paraphrase}. 

We then clarify the definition by assuming that combining the second and third components in Definition\,\ref{def:nlp} meet the definition of the paraphrase generation task~\cite{rus-etal-2014-paraphrase}. Henceforth, Definition\,\ref{def:nlp} can be simplified to Definition\,\ref{def:confined}.

\begin{definition}[Revise for concision at the sentence level, NLP, simplified]\label{def:confined}
Produce a \textit{paraphrase} where minimum wordiness can be identified.
\end{definition}

The revising\footnote{stands for \textit{(machine) revising for concision} if not otherwise specified, so does \textit{revision}} task is well-defined, as long as "paraphrase generation" is well-defined. It is a paraphrase generation task with a syntactic constraint.

\subsection{Task Performance Indicator}\label{sec:indicator}
How does one approximately measure revision performance? In principle, Definition~\ref{def:nlp} should be used as a checklist. A good sample requires correct grammar ($\gamma$), complete information ($\rho$) and reduced wordiness ($1-\omega$), assuming each component as a float number between 0 and 1. The overall assessment ($\chi$) of the three components is as follows,
\begin{equation}
    \chi = \alpha^2\cdot (\gamma - 1) + \alpha\cdot (\rho - 1) + (1-\omega),
\end{equation}
where $\alpha\in\mathbb{R}_{>1}$ is a large enough number, as we believe that $\gamma$ and $\rho$ overweigh $1-\omega$. Intuitively, if a revised sentence does not paraphrase the original one, assessing the reduction of wordiness makes little sense. Concision $\chi$ would always be negative if $\gamma<1$ or $\rho<1$.

Corresponding to the three components is a mix of three tasks, including grammatical error correction for $g$, textual semantic similarity for $r$, and wordiness detection for $w$. Unfortunately, both a reference-free metric good enough to characterize the paraphrase and a robust wordiness detector are rare. Therefore, such assessment of concision is now only feasible through human evaluation.

To enable automatic evaluation for faster feedback, we currently follow Papineni's viewpoint~\cite{papineni-etal-2002-bleu}. The closer a machine revision is to a professional human revision, the better it is. To judge the quality of a machine revision, one measures its closeness to one or more reference human revisions according to a numerical metric. Thus, our revising evaluation system requires two main components:
\begin{enumerate}
    \item A numerical "revision closeness" metric.
    \item A corpus of good quality human reference revisions.
\end{enumerate}

Different from days when Papineni needed to propose a closeness metric, we can adopt various metrics from machine translation and summarization community~\cite{lin-2004-rouge,banerjee-lavie-2005-meteor}. Since it is certain which criterion correlates best, we take multiple relevant and reasonable metrics into account to estimate quality of revision. These metrics include those measuring higher order n-grams precision (BLEU,~\citealp{papineni-etal-2002-bleu}), explicit Word-matching, stem-matching, or synonym-matching (METEOR,~\citealp{banerjee-lavie-2005-meteor}), surface bigram units overlapping (ROUGE-2-F1,~\citealp{lin-2004-rouge}), cosine similarity between matched contextual words embeddings (BERTScore-F1,~\citealp{zhang2019bertscore}), edit distance with single-word insertion, deletion, or replacement (word error rate,~\citealp{su1992new}), edit distance with block insertion, deletion, or replacement (translation edit rate,~\citealp{snover-etal-2006-study}), and explicit goodness of words editing against reference and source (SARI,~\citealp{xu-etal-2016-optimizing}). In short, BLEU, METEOR, ROUGE-2-F1, SARI, word error rate and translation edit rate estimate sentence wellformedness lexically; METEOR and BERTScore-F1 consider semantic equivalence. Comparing grammatical relations found in prediction with those found in references can also measure semantic similarity~\cite{clarke-lapata-2006-models,riezler-etal-2003-statistical,toutanova-etal-2016-dataset}. Grammatical relations are extracted from dependency parsing, and F1 scores can then be used to measure overlap.

In contrast, the lack of good parallel corpus impedes (machine) revising for concision. To address this limitation, we curate and make available such a corpus as benchmark. Each sample in the corpus contains a wordy sentence, and at least one sentence revised for concision. Samples are from English writing centres of 57 universities, ten colleges, four community colleges, and a postgraduate school.

\section{Related Work}
\label{sec:relatedWork}

Manual revision operations include delete, replace, and rewrite. Intuitively, a revising program should do similar jobs, too.  In fact, these actions are implemented individually in various NLP tasks. For example, sentence compression requires programs to delete unnecessary words, and paraphrasing itself is a matter of replacement. Machine revision for concision could also share traits with them. Practically, when a neural model learns in a Seq2Seq manner, the difference among these tasks is the parallel dataset. We are also interested in whether programs developed for these tasks can work in machine revision.

\subsection{Deleting as in Sentence Compression}\label{sec:sentcomp}
When revising, deleting redundant words is common. For example, we can revise "\textit{research is increasing in the field of nutrition and food science}" to "\textit{research is increasing in nutrition and food science}"~\cite{uri2019}, simply by deleting "\textit{the field of}". Deleting is canonical in sentence compression, a task aiming to reduce sentence length from source sentences while retaining basic meaning~\cite{jing-2000-sentence,knight2000statistics,mcdonald-2006-discriminative}. For example, the compression task has been formulated as integer linear programming optimization using syntactic trees~\cite{clarke-lapata-2006-constraint}, or as a sequence labelling optimization problem using the recurrent neural networks (RNN)~\cite{filippova-etal-2015-sentence,klerke-etal-2016-improving,kamigaito-etal-2018-higher}. They explicitly or implicitly use dependency grammar. Pre-trained language models such as ELMo~\cite{peters-etal-2018-deep} and BERT~\cite{devlin-etal-2019-bert} can encode features apart from dependency parsing~\cite{kamigaito2020syntactically}, bringing prediction and reference sentences closer.

All methods rely on parallel datasets labelling parts to be deleted. However, the deleting part in sentence compression differs from that in revision.~\citet{filippova-altun-2013-overcoming} created Google dataset from titles and first sentence of news articles. The information retained in the first sentence depends on the title. While this creation is useful for reducing excessive information, the deleted part is probably not wordiness.

Deleting does not solve everything in revision.  We can revise "\textit{in this report I will conduct a study of ants and the setup of their colonies}" to "\textit{in this report I will study ants and their colonies}", taking advantage of noun-and-verb homograph. However, a more concise version "\textit{this report studies ants}"~\cite{commnet2021} requires changing "\textit{study}" to third-person singular. 

\subsection{Replacing as in Paraphrase Generation}\label{subsec:paraphrase}
Word choice matters as well, thus we revise by paraphrasing to stronger words. Paraphrase generation changes a sentence grammatically and re-selects words, while retaining meaning. Paraphrasing matters in academic writing, for it helps avoid plagiarism. Rule-based or statistical machine paraphrasing substitutes words by finding synonyms from lexical databases, and decodes syntax according to template sentences. This rigid method may undermine creativity~\cite{bui2021generative}. Pre-trained neural language models like GPT~\cite{radford2019language} or  BART~\cite{lewis-etal-2020-bart} paraphrase more accurately~\cite{hegde2020unsupervised}. Through paraphrasing, we can replace verb phrase "\textit{conduct a study}" to verb "\textit{study}" in the example above, rather than delete and rely on noun-and-verb homographs to keep the sentence syntactically correct.

Machine revision is a kind of paraphrase generation, and vice versa is not true. Current paraphrase generation does not require concision in generated sentences. Automatically annotated datasets for paraphrasing include ParaNMT~\cite{wieting-gimpel-2018-paranmt}, Twitter~\cite{lan-etal-2017-continuously}, or re-purposed noisy datasets such as MSCOCO~\cite{lin2014microsoft} and WikiAnswers~\cite{fader-etal-2013-paraphrase}. We may adapt paraphrase parallel datasets to train revising models, as investigated in Section~\ref{sec:methods}.

\subsection{Other related tasks}

Summarization produces a shorter text of one or several documents, while retaining most of meaning~\cite{paulus2017deep}. This is similar to sentence compression. In practice, summarization welcomes novel words, allows specifying output length~\cite{kikuchi-etal-2016-controlling}, and removes much more information than sentence compression does. Datasets include XSum~\cite{narayan-etal-2018-dont}
, CNN/DM~\cite{hermann2015teaching}, WikiHow~\cite{koupaee2018wikihow}, NYT~\cite{sandhaus2008new}, DUC-2004~\cite{over2007duc}, and Gigaword~\cite{rush-etal-2015-neural}, where summaries are generally shorter than one-tenth of documents. On the other hand, sentence summarization~\cite{chopra-etal-2016-abstractive} uses summarization methods on sentence compression datasets, retaining more information and possibly generating new words.

Text simplification modifies vocabulary and syntax for easier reading, while retaining approximate meaning~\cite{omelianchuk-etal-2021-text}. Hand-crafted syntactic rules~\cite{siddharthan2006syntactic,carroll-etal-1999-simplifying,chandrasekar-etal-1996-motivations} and aligned
sentences-driven simplification~\cite{yatskar-etal-2010-sake} have been explored. Corpora such as Turk~\cite{xu-etal-2016-optimizing} and PWKP~\cite{zhu-etal-2010-monolingual} are compiled from Wikipedia and Simple English Wikipedia~\cite{coster-kauchak-2011-simple}. Rules for simplification may deviate from that for revision, \emph{e.g.}, text simplification sometimes encourages prepositional phrases~\cite{xu-etal-2016-optimizing}. Still, adapting these approaches may benefit academic revising for concision.

Fluency editing~\cite{napoles-etal-2017-jfleg} not only corrects grammatical errors but paraphrases text to be more native sounding as well. Its paraphrasing section is constrained such that outputs represent a higher level of English proficiency than inputs. As a constrained paraphrase task, fluency editing may alleviate ill-posed problems in paraphrase generation~\cite{cao-etal-2020-unsupervised-dual,rus-etal-2014-paraphrase}. However, such constraints may not be consistent with those required for concision.

In general, machine revision for academic writing requires new methods. Rules for revision can be adapted from these related tasks, so do training strategies.

\section{Benchmark Corpus}
\label{sectDataset}

The collated corpus, named \textbf{Concise-536}, contains 536 pairs of sentences. This is a fair starting size, comparable with 385 of RST-DT (semantic parsing,~\citealp{carlson2003building}), 500 of DUC 2004 (summarization\footnote{\url{https://duc.nist.gov/duc2004/}}), or 575 by~\citet{cohn-lapata-2008-sentence} (sentence compression). Each concise sentence is revised from its wordy counterpart by English specialists from the 72 universities, colleges, or community colleges. Sentence ID, category and original link are available for each data point\footnote{\url{https://huggingface.co/datasets}}, and a 120-point validation split from other sources is attached.

\begin{table*}[t]
\scriptsize
\centering
\begin{tabularx}{\textwidth}{lllXXX}
\hline
\textbf{Category} & \textbf{Action} & \textbf{\# sents.} & \textbf{Mean words wordy sent.} & \textbf{Mean words concise sent.} & \textbf{Translation Edit Rate} \\ \hline
\RomanNumeralCaps{1} & Delete & 169 & 13.16 & 9.17 & 4.72 \\
\RomanNumeralCaps{2} & Replace & 116 & 12.37 & 9.02 & 5.1 \\
\RomanNumeralCaps{3} & Rewrite & 153 & 14.43 & 9.73 & 9.54 \\
\RomanNumeralCaps{4} & Delete + Replace & 42 & 23.81 & 11.57 & 15.16 \\
\RomanNumeralCaps{5} & Replace + Rewrite & 33 & 21.52 & 12.85 & 14.88 \\
\RomanNumeralCaps{6} & Delete + Rewrite & 14 & 24.5 & 11.36 & 17.71 \\
\RomanNumeralCaps{7} & Delete + Replace + Rewrite & 9 & 32.56 & 14.56 & 25.56 \\
All & - & 536 & 15.32 & 9.86 & 8.31\\ \hline
\end{tabularx}
\caption{\label{tab:category}
Revising a sentence can involve either one of the three strategies (category \RomanNumeralCaps{1}, \RomanNumeralCaps{2}, \RomanNumeralCaps{3}), or a combination of them (category \RomanNumeralCaps{4}, \RomanNumeralCaps{5}, \RomanNumeralCaps{6}, \RomanNumeralCaps{7}). Sample sizes, average word counts before and after revisions, and average edit distance (translate edit rate, TER) for revision are listed.
}
\end{table*}

Revising different sentences can go through a completely different process. As seen below, simply crossing out a few words revises Example~\ref{ex:delete}; a new word is needed in revising Example~\ref{ex:replace}; and even the sentence structure needs changing in Example~\ref{ex:rewrite}.

\begin{corpus}[Delete]\label{ex:delete}
Any \remove{particular type of} dessert is fine with me.~\cite{purdue_writing_lab2021}
\end{corpus}

\begin{corpus}[Replace]\label{ex:replace}
She \replace{has the ability to}{can} influence the outcome.~\cite{purdue_writing_lab2021}
\end{corpus}

\begin{corpus}[Rewrite]\label{ex:rewrite}
\replace{The 1780 constitution of Massachusetts was written by John Adams.}{John Adams wrote the 1780 Massachusetts Constitution.}~\cite{north_carolina2021}
\end{corpus}

In Concise-536, we do not identify fine-grained wordiness because a phrase can have more than one type of verbosity at the same time. For instance, we can revise "\textit{Her poverty also helped in the formation of her character.}" to "\textit{Her poverty also helped form her character.}"~\cite{gmu2021}, treating "\textit{in the formation of}" as either a wordy prepositional phrase, or nominalization. Rather, we focus on editing. 

In editing, the three actions are not complementary, and instead have varying degrees of power. Deleting can be covered by replacing (See Section\,\ref{sec:sentcomp}), which could be again covered by rewriting, \emph{i.e.}, rewriting is the most flexible. However, Occam's razor pushes us to prioritize the actions requiring lower effort to complete, \emph{i.e.}, delete $<$ replace $<$ rewrite. Supposedly, the difficulty for implement each action with programs shares the same trend. In addition, some sentences contain multiple wordiness occurrences, each of which may need a different action, \emph{e.g.,} delete + replace.

Interested in how well a revising algorithm resembles each action, we label revisions in each sentence pair and divide them into seven categories. Revisions that require the same set of actions will be assigned to the same category. Each revision is assigned to one of seven categories in Table~\ref{tab:category}. 

For convenience, we standardize categorizing rules as follows where each sentence is a word-level sequence. For each pair, we have a wordy sequence ($w$) and a concise sequence ($c$).

\begin{enumerate}
    \item If $c$ is a (not necessarily consecutive) subsequence of $w$, we consider revision only requires deletion (category \RomanNumeralCaps{1}).
    \item If not, we only delete redundancy from $w$ to get $w'$, \emph{i.e.}, $w'$ paraphrases $w$, and $w'$ is a subsequence of $w$. Then, we make local\footnote{Empirically, in a sentence or clause, we do not replace the subject and predicate verb together.} replacement(s) to $w'$ to get $w^*$, and every individual state from $w'$ to $w^*$ (\emph{i.e.}, after each local replacement) paraphrases $w'$. If $w^* = c$ and $w' = w$, we consider revision only requires replacement (category \RomanNumeralCaps{2}). If $w^* = c$ and $w' \neq w$, we consider revision only requires deletion and replacement (category \RomanNumeralCaps{4}).
    \item If $w^* = w$, we consider revision relies solely on rewriting (category \RomanNumeralCaps{3}).
\end{enumerate}

\begin{corpus}[category \RomanNumeralCaps{1}]\label{exp:there}
\remove{There are} four rules \remove{that} should be observed.~\cite{purdue_writing_lab2021}
\end{corpus}

\begin{corpus}[category \RomanNumeralCaps{3}]\label{exp:regular}
\remove{Regular reviews of} online content should be \replace{scheduled}{reviewed regularly}.~\cite{monash2020}
\end{corpus}

\begin{corpus}[category \RomanNumeralCaps{4}]\label{exp:she}
She fell \remove{down} \replace{due to the fact that}{because} she hurried.
~\cite{purdue_writing_lab2021}
\end{corpus}

Example\,\ref{exp:there} used to be wordy in the running start, but deleting suffices in revision. Therefore, although counter intuitive, it belongs to category \RomanNumeralCaps{1}. An adjective-noun pair is the wordiness in Example\,\ref{exp:regular}, yet its revision is more complex than replacing a verb. Usually, revision involves multiple strategies, as seen in Example\,\ref{exp:she} (delete "\textit{down}" + replace "\textit{due to the fact that with}" with "\textit{because}").

Human annotators implement the rules, as we need to check whether the meaning is still the same at each step.

Usually, category \RomanNumeralCaps{3} sentences are the hardest to revise as the easier strategies of deleting and replacing are not applicable. In fact, revising category \RomanNumeralCaps{5}, \RomanNumeralCaps{6}, and \RomanNumeralCaps{7} sentences are more challenging, as these sentences are longer, more complex, and more deliberate than category \RomanNumeralCaps{3} sentences (Figure~\ref{fig:difficulty}, Table~\ref{tab:good},\,\ref{tab:bad}), which is a bias in this corpus.

\section{Approaches to Revisions}
\label{sec:methods}

We approach the raised problem in this study. Solutions to machine revision for concision can be diverse. Neural model solutions include tree-to-tree transduction models~\cite{cohn-lapata-2008-sentence}, or general Seq2Seq models. We present a Seq2Seq approach, for it is flexible and straightforward. The model architecture is BART~\cite{lewis-etal-2020-bart}.

Ideally, training corpora tune statistical models or neural models, such that we can test tuned models on the benchmark corpus. However, lacking authoritative revisions prompted us to let models fit relevant task data. We also use public external knowledge, \emph{e.g.}, WordNet~\cite{fellbaum2010wordnet}. This section describes how we build an ad hoc training corpus to initiate this task.

The BART base model (124,058,116 parameters) is then used to fit \emph{each} training set in this section. Training settings are fixed (batch size at 32, PyTorch Adam optimizer~\cite{NEURIPS2019_9015,kingma2014adam} with initial learning rate at $5\times10^{-5}$, validated every 5,000 iterations). We then evaluate trained models on Concise-536.

\subsection{Baselines}
We prepare training samples by adjusting data from paraphrase generation (ParaNMT,~\citealp{wieting-gimpel-2018-paranmt}), sentence simplification (WikiSmall,~\citealp{zhang-lapata-2017-sentence}), or sentence compression (Gigaword,~\citealp{rush-etal-2015-neural}; Google News datasets,~\citealp{filippova-altun-2013-overcoming}; MSR Abstractive Text Compression Dataset,~\citealp{toutanova-etal-2016-dataset}).

\subsection{Approach 1: WordNet as Booster}

Baseline methods are useful, but they are not developed for revision tasks after all. 
To replace a verb or noun phrase with a single word, we leverage word glosses in public dictionaries, \emph{i.e.}, WordNet~\cite{fellbaum2010wordnet}. Word semantics are close to semantics in their glosses. This feature is usually used to improve word embedding~\cite{bosc-vincent-2018-auto} or evaluate analogy of word embedding~\cite{mikolov2013efficient}. We use this feature to replace a verb or noun phrase with a single word.

We create data samples using WordNet and a language modelling corpus. For each sentence $s$ in the corpus, we use WordNet vocabulary glosses to inflate it and obtain $s'$. Resulted parallel data approximate phrase replacement in sentence revision.

We first pick a unigram $u$, one of nouns, verbs, adjectives, or adverbs in $s$. At the same time, we avoid common words, \emph{e.g.}, "\textit{old}", or collocations and compounds, \emph{e.g.}, "\textit{united}" in "\textit{United Kingdom}". Next, we apply Lesk's dictionary-based word sense disambiguation (WSD) algorithm~\cite{lesk1986automatic} on $u$ and $s$ to get gloss $g$. Then, we parse $s$ and $g$ to obtain respective dependency trees $T_s$ and $T_g$; $r_g$ denotes root node in $T_g$. Usually, if $u$ is a noun, $r_g$ is a noun, and if $u$ is an adjective, $r_g$ is a verb. Eight $u \rightarrow r_g$ patterns account for over 90\% of the WordNet vocabulary (Table~\ref{tab:pos}). In Algorithm~\ref{alg:tree}, we modify dependency trees ($T_s$ and $T_g$) according to the eight patterns. The remaining six patterns are NOUN$\rightarrow$VERB, ADJ(-S) $\rightarrow$ ADJ, ADJ $\rightarrow$ ADP, ADJ $\rightarrow$ VERB, and ADV $\rightarrow$ ADP.% For details, visit \url{https://github.com/cestwc/dictionary-based-sentence-expander}

\begin{algorithm}[t!]
\footnotesize
\begin{algorithmic}
\Require $T_s, T_g$
\Return $s'$
% \State $\hat{\mathbf{s}} \gets ()$
\State Copy-children(from $u$, to $r_g$)
\State Locate $h_u$ \Comment{head node of $u$}
\State Delete ($u$ with children, from $T_s$)
\If{$u \in$  NOUN}
\State Insert-child-node ($r_g$ with children, to $h_u$)
\If {$r_g \in$  VERB}
\State $u \gets$ Gerund($u$)
\EndIf
\State Correct inflections (singular and plural forms)
\State Remove duplicate determiners
\ElsIf{$u \in$  VERB}
\State Insert-child-node ($r_g$ with children, to $h_u$)
\State Correct inflections (person and tense)
\State Add/Remove prepositions according to verb transitivity
\Else
\State Insert-right-child-node ($r_g$ with children, to $h_u$) \Comment{Post attributive}
\EndIf
\State $s'\gets$ Linearize($T_s$)
\end{algorithmic}
\caption{Rule-based Gloss Substitution}\label{alg:tree}
\end{algorithm}

Finally, we filter and post-process synthesized sentences. We parse $s'$ again and compare it with the dependency tree from which $s'$ is linearized. We drop those with more than three mismatches, or with accuracy lower than 0.9. We "smooth" synthesized sentences with parroting\footnote{\url{https://huggingface.co/prithivida/parrot_paraphraser_on_T5}}, to mitigate overfitting. We also drop those sharing low semantic similarity (BERTScore $\leq$ 0.82) with original $s$.

We take the first 0.2 million sentences from WikiText-103 corpus~\cite{merity2016pointer} and around 71 thousand data points after filtration are available to train the BART base model.

\begin{table*}[t!]
\small
\centering{
\begin{tabularx}{\textwidth}{Xlllllllllll}
\hline
\textbf{Methods} & \textbf{\RomanNumeralCaps{1}}& \textbf{\RomanNumeralCaps{2}}& \textbf{\RomanNumeralCaps{3}}& \textbf{\RomanNumeralCaps{4}}& \textbf{\RomanNumeralCaps{5}}& \textbf{\RomanNumeralCaps{6}}& \textbf{\RomanNumeralCaps{7}} & \textbf{All}   & \textbf{H} & $\rho$ & $\omega$  \\
\hline
ParaNMT~\cite{wieting-gimpel-2018-paranmt}   & 0.46          & \textbf{0.62} & 0.46          & \textit{0.53} & 0.44          & 0.45          & \textit{0.38} & 0.55   &    3.40  & &  \\
MSR~\cite{toutanova-etal-2016-dataset}       & \textit{0.74} & 0.58          & 0.44          & 0.51          & 0.41          & 0.44          & 0.37          & 0.57     &   \textit{2.79} & 0.78 & \textbf{0.40} \\
G. News~\cite{filippova-altun-2013-overcoming}      & \textit{0.61}          & 0.46          & 0.39          & 0.40          & 0.35          & 0.39          & 0.33          & 0.48      &  5.74 & & \\
Gigaword~\cite{rush-etal-2015-neural}  & 0.30          & 0.29          & 0.28          & 0.31          & 0.25          & 0.29          & 0.23          & 0.29    &  6.74  & &  \\
WikiSmall~\cite{zhang-lapata-2017-sentence} & 0.70          & 0.59          & \textbf{0.48} & 0.52          & 0.44          & \textbf{0.46} & \textit{0.38} & 0.57      &   3.31 & &\\
Our Approach 1   & 0.70          & 0.60          & \textit{0.47}          & 0.52          & \textbf{0.46} & \textit{0.45}          & 0.37          & \textit{0.58}    &    \textit{2.79}&\textbf{0.99}&0.47\\
Our Approach 2  & \textbf{0.75} & 0.60          & \textit{0.47 }         & \textbf{0.55} & 0.40          & \textit{0.45}          & \textbf{0.41} & \textbf{0.59} &  \textbf{2.62} &\textit{0.82}&\textit{0.41}\\
% Input (control group)     & 0.73          & \textit{0.61} & \textbf{0.49} & \textit{0.53} & \textbf{0.46} & \textbf{0.47} & \textit{0.38} & \textbf{0.59}  & -& - &0.52\\
\hline
\end{tabularx}
}
\caption{We average BLEU, METEOR, ROUGE-2-F1, SARI, Parsed relation F1, BERTScore-F1, \textit{and} (negative) translation edit rate of (pre-)baseline methods. The most favorable score in each column is in bold, the second most favorable in italics. This table estimates the strengths and weaknesses of each variants.
% Detailed statistics of different categories on each metric are shown in Table~\ref{tab:results}.
System ranking from human evaluation (\textbf{H}), information retention ($\rho$), and wordiness ($\omega$) are presented in the right-most columns.  }\label{tab:brief_results}
\end{table*}

\subsection{Approach 2: Multi-Task Learning}\label{sec:mix}

Each dataset in baselines and Approach 1 handles part of task. However, sentence compression or simplification does not emphasize complete information retention; paraphrase generation hardly encourages deletion; synthetic data limit editing scope because word glosses are limited. We hypothesize that mixing the good samples among these datasets could more closely approximate the revision task. Therefore, we adjust datasets again. We keep every sample in MSR as it is small (21,145, see Appendix). Semantic similarity lower bound for sentence compression and simplification datasets is set at BERTScore = 0.9.  For ParaNMT, we discard samples with less than 10 words. As a result, ablation of mixed and shuffled data samples shows that a mixture of MSR, filtered ParaNMT, and synthetic WordNet dataset leads to the strongest approach. This approach uses transfer learning from multiple datasets to learn revising strategies such as deletion and phrase replacement.

\subsection{Experiment and Result}
\begin{table*}[!t]
\centering
\small
\begin{tabularx}{\textwidth}{lXX}
\hline
\textbf{Category} & \textbf{Reference(s)} & \textbf{Prediction} \\\hline
5th percentile & Bob \replace{provided an explanation of}{explained} the computer to his grandmother. & Bob \replace{provided an explanation of}{explained} the computer to his grandmother. \\
95th percentile & \remove{Rather than taking the bull by the horns,} she  \replace{was quiet as a church mouse}{avoided confrontation by remaining silent}. & \remove{Rather than taking the bull by the horns,} she was quiet as a church mouse. \\
\hline
\end{tabularx}
\caption{\label{tab:illustrate}
Well/poorly revised samples in the corpus. Shorter sentences that require simpler actions are perfectly revised. Rewriting clich\'es is difficult, in which case the approach tends to use deletion.
% Table~\ref{tab:good} and \ref{tab:bad} show sample machine revision for sentences from seven categories.
}
\end{table*}
Table~\ref{tab:brief_results} shows test results in each category. Our approach 2 has the highest overall score and is more robust than baseline models on category \RomanNumeralCaps{1}, \RomanNumeralCaps{4}, and \RomanNumeralCaps{7}. The same architecture trained only on MSR outperforms any other baseline for deletion (category \RomanNumeralCaps{1}) and ranks second for replacement (category \RomanNumeralCaps{2}). The top-ranked baseline for replacement (category \RomanNumeralCaps{2}) is trained on ParaNMT. In category \RomanNumeralCaps{5}, the model trained on WordNet scores highest, slightly outperforming other baselines. Trends in category \RomanNumeralCaps{3}, \RomanNumeralCaps{4}, \RomanNumeralCaps{6}, \RomanNumeralCaps{7} are less clear. Datasets Gigaword, Google News, and WikiSmall may be quite different from the benchmark corpus, and thus models trained on these datasets do not score well.

Our approach 2 suffers from two shortcomings common to all baselines. First, the model relies on transfer learning from MSR and ParaNMT and struggle to rewrite (category \RomanNumeralCaps{3}) or to handle composite wordiness (category \RomanNumeralCaps{5}, \RomanNumeralCaps{6}, \RomanNumeralCaps{7}). Second, the approach 2 outputs score worse than the input text on many metrics in many categories, especially on category \RomanNumeralCaps{3}. These shortcomings suggest challenges in revision. We take the 5th and 95th percentile from all 536 samples to qualitatively illustrate the best proposed approach in Table~\ref{tab:illustrate}. Apart from samples in Concise-536, Figure~\ref{fig:obscure} shows an arbitrary sentence by non-English native speakers~\cite{chen2020survey}. The proposed revisioner removes repetition and unnecessary prepositional phrases, illustrating its potential in academic writing.

For human evaluation, we adopt an approach similar to \citet{hsu-etal-2018-unified,zhang2020pegasus,ravaut2022summaReranker}. We (1) rank the samples by overall automatic evaluation on the model in descending order; (2) divide the examples in \textit{each category} into two buckets; (3) randomly pick one example from each bucket. For each picked sample, we ask three graduate students (IELTS 7.0 or equivalent) to rank the predictions of seven systems, and the average ranking of each system is shown in \textbf{H} column in Table~\ref{tab:brief_results}.

For top three systems, human evaluators then assess information retention ($\rho$) and wordiness ($\omega$), since system outputs are in good syntax. Particularly, human assessment on wordiness engages the Paramedic Method~\cite{lanham1992revising} to highlight the wordy part and $\omega = $ (\# wordy words) / (\# all words). The model trained adapted WordNet data preserves information better, which also accounts for its good human ranking.

We observe general correlation between automatic score ranking and human evaluation ranking. However, information retention is not sufficiently represented by semantic similarity scores like BERTScore. These findings suggest further investigation on the evaluation scheme of this task.

% \begin{example}
% Finding the attack features and attack methods of \remove{the} adversarial examples is \remove{the core of} our problem solving.
% \end{example}

\section{Discussion}
\label{sec:discussion}

Comparing the proposed revisioner's effectiveness for different categories, we understand deleting and replacing are much easier sub-tasks than rewriting is. The former two actions, especially deletion, are less ill-posed, while rewriting is open. Still, revision for concision requires an algorithm that is able to use all three actions in combination. Its goal is to resolve all seven categories of cases, marking distinction between revision and other tasks such as sentence compression.

We use seven metrics to estimate a revisioner's effectiveness, since each metric has its shortcomings. For example, METEOR does not adequately penalize nominalization, and thus wordy input texts typically score higher on METEOR than algorithm outputs. More targeted metrics for this task, including reference-free structural metrics~\cite{sulem-etal-2018-semantic}, might help. We do not include word counts. Although concision is marked by brevity and wordiness often correlates to high word count, concise writing does not always require the fewest words~\cite{purdue_writing_lab2021}. Optimizing a lower word count may be misleading even if it is constrained to zero information loss~\cite{siddharthan2006syntactic}. For example, abusing pronouns and ellipses can result in shorter sentences that are harder to read. 

Transferring knowledge from other tasks to approximate revising is a stopgap measure. Specialized revising methods exist, \emph{e.g.}, the Paramedic Method~\cite{lanham1992revising}. Automated specialized methods may be more efficient.

\section{Conclusion}

We formulate sentence-level revision for concision as a constrained paraphrase generation task. The revision task not only requires semantics preservation as in usual paraphrasing tasks, but also specifies syntactic changes. A revised sentence is free of wordiness and as informative. Revising sentences is challenging and requires coordinated use of delete, replace, and rewrite. To benchmark revising algorithms, we collect 536 sentence pairs before and after revising from 72 college writing centres. We then propose a Seq2Seq revising model and evaluate it on this benchmark. Despite scarcity of training data, the proposed approaches offer promising results for revising academic texts. We believe this corpus will drive specialized revision algorithms that benefit both authors and readers.

% \begin{table}
% \centering
% \begin{tabular}{lc}
% \hline
% \textbf{Command} & \textbf{Output}\\
% \hline
% \verb|{\"a}| & {\"a} \\
% \verb|{\^e}| & {\^e} \\
% \verb|{\`i}| & {\`i} \\ 
% \verb|{\.I}| & {\.I} \\ 
% \verb|{\o}| & {\o} \\
% \verb|{\'u}| & {\'u}  \\ 
% \verb|{\aa}| & {\aa}  \\\hline
% \end{tabular}
% \begin{tabular}{lc}
% \hline
% \textbf{Command} & \textbf{Output}\\
% \hline
% \verb|{\c c}| & {\c c} \\ 
% \verb|{\u g}| & {\u g} \\ 
% \verb|{\l}| & {\l} \\ 
% \verb|{\~n}| & {\~n} \\ 
% \verb|{\H o}| & {\H o} \\ 
% \verb|{\v r}| & {\v r} \\ 
% \verb|{\ss}| & {\ss} \\
% \hline
% \end{tabular}
% \caption{Example commands for accented characters, to be used in, \emph{e.g.}, Bib\TeX{} entries.}
% \label{tab:accents}
% \end{table}

% \begin{table*}
% \centering
% \begin{tabular}{lll}
% \hline
% \textbf{Output} & \textbf{natbib command} & \textbf{Old ACL-style command}\\
% \hline
%~\citep{Gusfield:97} & \verb|~\citep| & \verb|~\cite| \\
%~\citealp{Gusfield:97} & \verb|~\citealp| & no equivalent \\
%~\citet{Gusfield:97} & \verb|~\citet| & \verb|\newcite| \\
%~\citeyearpar{Gusfield:97} & \verb|~\citeyearpar| & \verb|\shortcite| \\
% \hline
% \end{tabular}
% \caption{\label{citation-guide}
% Citation commands supported by the style file.
% The style is based on the natbib package and supports all natbib citation commands.
% It also supports commands defined in previous ACL style files for compatibility.
% }
% \end{table*}

\section*{Ethical considerations}
The release of Concise-536 is intended only for "not-for-profit" educational purposes or private research and study in accordance with the Copyright Act 1994; all original text content is acknowledged as the property of each educational institution. All text content in Concise-536 (and the 120-point validation split) are public, and our release details their original links, thus making the release no different from a list of outbound links.

\section*{Limitations}
The transfer of knowledge from other tasks to the rough revision is an emergency solution. There are specialised revision methods. For example, automating the Paramedic method~\cite{lanham1992revising} could possibly lead to a more efficient revisioner.

\section*{Acknowledgements}

We would like to thank the reviewers for thoughtful comments and efforts to improve our manuscript.

\bibliography{anthology,custom,sac}
\bibliographystyle{acl_natbib}

\appendix

\section{Linguistic Rules in Revising for Concision}
We collate and present a set of practical linguistic rules for concise sentence revision, which we synthesize based on guidelines from writing centres at numerous major universities and educational institutes. Table~\ref{tab:strategies} illustrates how wordiness can be fine-grained, and what action is required once a wordiness is identified~\cite{north_carolina2021,purdue_writing_lab2021,monash2020}.

\begin{table}[!t]
\centering
\small
\begin{tabularx}{0.5\textwidth}{Xl}
\hline
\textbf{Wordiness identified} &   \textbf{Action}        \\ \hline
           Weak modifiers                         &Delete\\
                 \,\,(qualifiers / intensifiers)            &\\
                 Redundant pairs                        &\\
                 Grouped synonyms                       &\\
                 Stock phrases                          &\\
                 Unnecessary hedging                    &\\
                 Implied information                    &\\
                 Yourself                               &\\ \hline
          Informal language                      &Replace\\
                 Vague pronoun references               &\\
                 Possessive constructions using "of"    \\
                 Prepositional phrases                  &\\
                 All-purpose nouns                      &\\
                 Vague Swamp                            &\\
                 Fancy words                            &\\
                 Helping verbs                          &\\
                 \,\,("to be" verbs, "be" + adjective)      \\
                 Adjective-noun pairs                   &\\
                 Phrasal verbs                          &\\
                 Verb-adverb pairs                      &\\
                 Nominalisation / noun strings          \\
                 Cliches and Euphemisms                 &\\
                 Empty phrases                          &\\
                 Expletive constructions                &\\ \hline
          long sentences (\textgreater 25 words) &Rewrite\\
                 Running starts                         &\\
                 \,\,(with "there / it" + "be")             &\\
                 Long opening phrases / clauses         \\
                 Needless transitions                   &\\
                 Interrupted subjects and verbs         \\
                 Interrupted verbs and objects          \\
                 Negatives (opposite to affirmatives)   \\
                                                        &\\
                 \textit{and anything violating:}       \\
                 A blend of active and passive verbs    \\
                 Elliptical constructions / parallelism \\
                 Only one main idea per sentence        \\ \hline
\end{tabularx}

\caption{\label{tab:strategies}
Revising rules collated from college writing centers. Three actions are available. Redundancy can be deleted; short, specific, concrete and stronger expressions shall replace vague ones; sentences should be rewritten if neither deleting nor replacing helps.
}
\end{table}

\section{Technical Difficulties in Reference-free Revision Evaluation}

Had we chosen not to follow Papineni's viewpoint, reference-free evaluation is the way to go. However, it is technically not trivial to use programs to detect wordiness or syntax errors these days (See Section~\ref{sec:indicator}), let alone detect semantic similarity. Progress in sentence embedding~\cite{lin2017structured} and semantic textual similarity~\cite{yang2019xlnet} enables meaning comparison between sentences, but relying on one developing system to evaluate another is risky. Moreover, information delivered by a sentence is sometimes beyond its textual meaning. Concise writing can suggest eliminating  first-person narratives; \emph{e.g.}, "\textit{I feel that the study is significant}" is revised to "\textit{The study is significant}"~\cite{waldenu2021}. Here, the first-person statement used to be the main clause, and removing it will shift sentence embedding. Nevertheless, in academic writing, these two sentences deliver identical information.

\section{Balance between Syntax, Information, and Wordiness}
The coefficient $\alpha$ tells how much syntax overweighs information, or information overweighs reduced wordiness. Empirically, minimum of $\alpha$ can be around the word count in a standard sentence. In other words, even if a single key word is missing, the decrease in $\rho$ is bigger than the increase in $1 - \omega$. 

\section{Explaining Categories in the Corpus}
\begin{figure}[h!]
    \centering
    \small
    \begin{tikzpicture}
        \begin{scope}[blend group=soft light]
        \fill[p13!70!white]   ( 209.98945070661557:1.1982899710040975) circle (1.4911131108882707);
		
        \fill[b14!70!white] ( 95.9701298441318:1.020383701219179) circle (1.3785345586728925);
        		
        \fill[y15!70!white]  ( -13.379609620139817:1.2988177887329182) circle (1.409210281782305);
        		
        \end{scope}
        \node at ( 209.98945070661557:1.6776059594057364)    {\RomanNumeralCaps{1} };
        		
        \node at ( 95.9701298441318:1.4285371817068504)    {\RomanNumeralCaps{2} };
        		
        \node at ( -13.379609620139817:1.8183449042260853)    {\RomanNumeralCaps{3} };
        		
        \node at ( 152.9797902753737:0.679315988401639)    {\RomanNumeralCaps{4} };
        		
        \node at ( 41.295260111995994:0.679315988401639)     {\RomanNumeralCaps{5} };
        		
        \node at ( -82.75:0.679315988401639)     {\RomanNumeralCaps{6} };

        \node at ( 209.98945070661557:3.275895930409834)    {Delete};
        		
        \node at ( 95.9701298441318:2.9489208829260294)    {Replace};
        		
        \node at ( -13.379609620139817:3.3171626929590037)    {Rewrite};
        		
        \node at (0:0)   {\RomanNumeralCaps{7} };
    \end{tikzpicture}
    \caption{Revising a sentence can involve either one of the three strategies (category \RomanNumeralCaps{1}, \RomanNumeralCaps{2}, \RomanNumeralCaps{3}), or a combination of them (category \RomanNumeralCaps{4}, \RomanNumeralCaps{5}, \RomanNumeralCaps{6}, \RomanNumeralCaps{7}).}
    \label{fig:categories}
\end{figure}
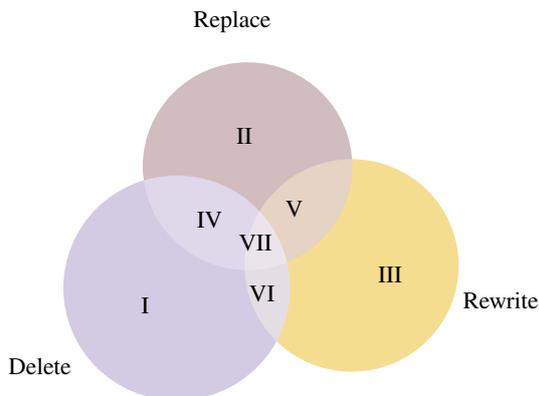
There are seven categories, as seen in Figure~\ref{fig:categories}. Note that although three actions (delete, replace, rewrite) are put side-by-side, they are with different levels of flexibility. In fact, every revision made with deleting can be done through replace, \emph{e.g.}, the fourth example in Table~\ref{tab:good}, "\textit{fell down}" could be replaced to "\textit{fell}", but we would simply consider the cheapest revision, which is to delete "\textit{down}"\footnote{"\textit{fell}" means "\textit{fell down}", as one never "\textit{fell up}".}. Similarly, rewriting is even more expensive and ambiguous. Therefore, our rule of Occam's razor is that only when a cheaper revision fails, will we use a more expensive one.

Here, we give examples of which category a revision corresponds to. Indeed, many sentence revisions are categorized in original websites. See the two examples from Purdue Writing Lab~\cite{purdue_writing_lab2021} below. The strategies applied are to "eliminate words that explain the obvious or provide excessive detail" (category \RomanNumeralCaps{1}) and to "replace several vague words with more powerful and specific words" (category \RomanNumeralCaps{2}), respectively.

\begin{corpus}[category \RomanNumeralCaps{1}]
Imagine \remove{a mental picture of} someone \remove{engaged in the intellectual activity of} trying to learn \remove{what} the rules \remove{are for how to play the game} of chess.

\end{corpus}

\begin{corpus}[category \RomanNumeralCaps{2}]
The politician \replace{talked about several of the merits of}{touted} after-school programs in his speech
\end{corpus}

For revisions not categorized in sources, we first align the segments of a pair of sentences by their meaning, as seen in Figure~\ref{fig:obscure}. This is intuitively straightforward when the revised sentence is given\footnote{If the revised sentences were not from a trustworthy site, this process could have been less intuitive.}.

Then, we determine the actions to revise. For example, in the fourth example (category \RomanNumeralCaps{4}) in Table~\ref{tab:good}, we find that we cannot delete any words in "\textit{due to the fact that}" without violating the second and third components in Definition\,\ref{def:nlp}. Thus, we have to put some more concise conjunction to take its place, \emph{i.e.}, "\textit{because}".

Another example is the sixth one (category \RomanNumeralCaps{4}) in Table~\ref{tab:bad}. Though it looks that the entire wordy sentence can only be written to reach the concise form, a cheaper revision is actually to first delete some redundancy, \emph{e.g.}, "\textit{sent \remove{to you} by us}", and then rewrite the necessary part.

Whether the subject and predicate of a sentence (clause) is changed together determines the border between replacing and rewriting. In the fifth example (category \RomanNumeralCaps{4}) in Table~\ref{tab:good}, "\textit{it was necessary}" is aligned to "\textit{had to}", and "\textit{us}" to "\textit{we}". However, we cannot change either of them individually without violating the third component in Definition\,\ref{def:nlp}. Therefore, when two or more replacements intertwine, we rewrite.

\section{Explaining Rule-based Gloss Substitution}
\label{sec:graft}

\begin{figure*}[!t]
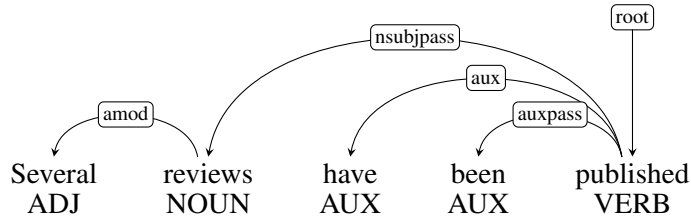
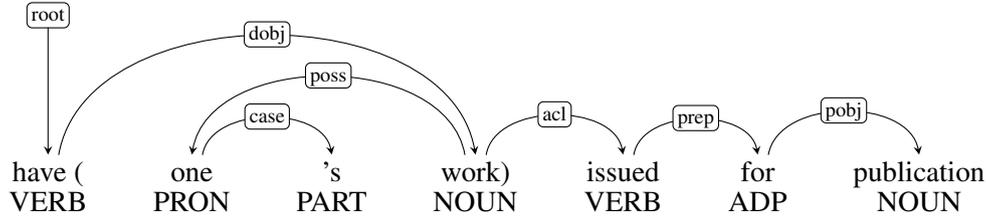
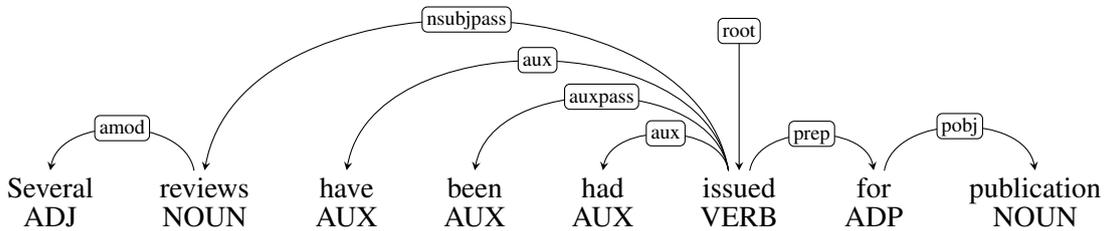

    \centering
    \begin{subfigure}{\textwidth}
    \centering
    \begin{dependency}[arc edge, arc angle=80]
    \begin{deptext}[column sep=.7cm]
    Several \& reviews \& have \& been \& published \\
    ADJ \& NOUN \& AUX \& AUX \& VERB \\
    \end{deptext}
    \deproot{5}{root}
    \depedge{2}{1}{amod}
    \depedge{5}{2}{nsubjpass}
    \depedge{5}{3}{aux}
    \depedge{5}{4}{auxpass}
    \end{dependency}
    \caption{Sentence "\textit{Several reviews have been published}" and its dependency tree. We expand the word "\textit{publish}" below.}
    \label{fig:s1}
    \end{subfigure}\\
    \begin{subfigure}{\textwidth}
    \centering
    \begin{dependency}[arc edge, arc angle=80]
    \begin{deptext}[column sep=.7cm]
    have ( \& one \& 's \& work) \& issued \& for \& publication \\
    VERB \& PRON \& PART \& NOUN \& VERB \& ADP \& NOUN \\
    \end{deptext}
    \deproot{1}{root}
    \depedge{1}{4}{dobj}
    \depedge{4}{2}{poss}
    \depedge{2}{3}{case}
    \depedge{4}{5}{acl}
    \depedge{5}{6}{prep}
    \depedge{6}{7}{pobj}
    \end{dependency}
    \caption{Gloss of "\textit{publish}" from WordNet\cite{fellbaum2010wordnet}; the root node is the verb "\textit{have}".}
    \label{fig:g1}
    \end{subfigure}\\
    \begin{subfigure}{\textwidth}
    \centering
    \begin{dependency}[arc edge, arc angle=80]
    \begin{deptext}[column sep=.7cm]
    Several \& reviews \& have \& been \& had \& issued \& for \& publication \\
    ADJ \& NOUN \& AUX \& AUX \& AUX \& VERB \& ADP \& NOUN \\
    \end{deptext}
    \deproot{6}{root}
    \depedge{2}{1}{amod}
    \depedge{6}{2}{nsubjpass}
    \depedge{6}{3}{aux}
    \depedge{6}{4}{auxpass}
    \depedge{6}{5}{aux}
    \depedge{6}{7}{prep}
    \depedge{7}{8}{pobj}
    \end{dependency}
    \caption{Synthesized sentence with the first stage of our approach; $r_g$ node "\textit{have/had}" is grafted onto the original sentence in (a).}
    \label{fig:ss1}
    \end{subfigure}\\

    \caption{Demonstration of dependency tree grafting in sentence synthesis. The dependency in (c) is obtained by re-parsing the synthesized sentence. As we can see, the POS tag of "\textit{have/had}" has changed from a verb to an auxiliary word, and the synthesized sentence is still syntactically and semantically correct, which shows that dependency changes may be unavoidable in the process of sentence synthesis. We also dealt with inflections to reduce grammatical errors.}
    \label{fig:graft}
\end{figure*}

A demonstration of Algorithm~\ref{alg:tree} is shown in Fig.\ref{fig:graft}, where a verb that appears in the past participle is replaced. By running this rule-based gloss replacement multiple times, we can recursively expand a sentence because the words used in a gloss have their associated glosses~\cite{bosc-vincent-2018-auto}. Table~\ref{tab:pos} describes $u \rightarrow r_g$ in the WordNet vocabulary.

\begin{table}[h!]
\centering
\small
\begin{tabularx}{.5\textwidth}{lXXXXX}
\hline
\textbf{POS} & \textbf{ADJ} & \textbf{ADJS} & \textbf{ADV} & \textbf{NOUN} & \textbf{VERB} \\ \hline
VERB & \textbf{3627} & \textbf{6354} & 221 & \textbf{3349} & \textbf{11586} \\
DET & 1 & 6 & 4 & 1594 & 0 \\
ADJ & 1053 & \textbf{2825} & 50 & 544 & 316 \\
NOUN & 155 & 405 & 57 & \textbf{73527} & 1739 \\
CCONJ & 0 & 0 & 0 & 24 & 0 \\
PUNCT & 0 & 1 & 0 & 6 & 0 \\
PART & 0 & 5 & 4 & 4 & 6 \\
ADV & 10 & 84 & 235 & 37 & 52 \\
ADP & \textbf{2615} & 972 & \textbf{3019} & 222 & 29 \\
AUX & 5 & 5 & 4 & 108 & 1 \\
PRON & 0 & 3 & 2 & 1516 & 1 \\
SCONJ & 4 & 10 & 14 & 3 & 10 \\
PROPN & 0 & 6 & 0 & 534 & 15 \\
X & 2 & 2 & 2 & 16 & 19 \\
NUM & 1 & 16 & 8 & 658 & 0 \\
INTJ & 1 & 1 & 3 & 13 & 12 \\
SYM & 0 & 0 & 0 & 0 & 0 \\ \hline
\end{tabularx}
\caption{Part-of-speech (POS) tags for a word $w$ and its corresponding $r_g$. Representation of POS tags follows the Stanford typed dependencies manual~\cite{de2008stanford} (except for ADJ-S, which stands for 'adjective satellite' in WordNet~\cite{fellbaum2010wordnet}). POS tags of $r_g$ are closely related to the POS tags of $w$, and we bold the pairs that appear frequently. In particular,  among nearly 117,000 word-gloss ($w \rightarrow r_g$) pairs, NOUN $\rightarrow$ NOUN is most frequent, accounting for more than three fifths. We have now studied the eight most frequently occurring pairs.}
\label{tab:pos}
\end{table}

% http://ctan.math.illinois.edu/graphics/pgf/contrib/tikz-dependency/tikz-dependency-doc.pdf

\section{Details in Datasets Used to Train baselines}

We prepare training samples by adjusting data from paraphrase generation, sentence simplification, or sentence compression. ParaNMT~\cite{wieting-gimpel-2018-paranmt} contains over five million paraphrase pairs annotated from machine translation tasks; we sort each pair by sentence length. This is a rough approximation, since shorter sentences are not necessarily more concise. Google News datasets (News,~\citealp{filippova-altun-2013-overcoming}) contains 0.2 million pairs of sentences, where the longer one is the leading sentence of each article, and the shorter one is a subsequence of the longer one. Gigaword~\cite{rush-etal-2015-neural} contains four million pairs of article headline and the first sentence. Although these datasets are mainly for generating news headlines~\cite{p-v-s-meyer-2019-data}, they approximate the deletion aspect of sentence revision. MSR Abstractive Text Compression Dataset~\cite{toutanova-etal-2016-dataset} contains six thousand sentence pairs from business letters, newswire, journals, and technical documents sampled from the Open American National Corpus\footnote{\url{https://www.anc.org/data/oanc}}; humans rewrite sentences at a fixed compression ratio.  WikiSmall~\cite{zhang-lapata-2017-sentence} contains sentence pairs from Wikipedia articles and corresponding Simple English Wikipedia. We adopt training splits of these datasets, and Table~\ref{tab:trainsize} lists their sizes.

\begin{table}[h!]
\small
\centering
\begin{tabular}{ll}
\hline
\textbf{Dataset} & \textbf{Size}\\
\hline
MSR (~\citeyear{toutanova-etal-2016-dataset}) & 21,145 \\
ParaNMT (~\citeyear{wieting-gimpel-2018-paranmt}) & 5,306,522 \\
Google News (G. News~\citeyear{filippova-altun-2013-overcoming}) & 200,000 \\ 
Gigaword (~\citeyear{rush-etal-2015-neural}) & 3,803,957 \\ 
WikiSmall (~\citeyear{zhang-lapata-2017-sentence}) & 89,042 \\\hline
Approach 2 fine-tuning set (Section~\ref{sec:mix}) & 182,330 \\\hline
\end{tabular}
\caption{Sample numbers of training sets. MSR dataset has multiple references, we take each reference as a sample point. The mixed fine-tuning set in Section~\ref{sec:mix} is composed of 89,712 samples from ParaNMT, 21,145 from MSR, and 71,473 from our synthesized dataset from WordNet.}
\label{tab:trainsize}
\end{table}

% All 124,058,116 parameters are trained as a whole on the composed train set for one epoch. When the batch size is eight by 2 GPUs, the training time on two RTX 2080Ti graphics processing units (GPU) is less than an hour. Learning rate starts from 5e-5 and PyTorch Adam optimizer is used.
\section{Random Sample Selection in Human Evaluation}
\small
\begin{lstlisting}
import random
random.seed(0)
for k in [169, 116, 153, 42, 33, 14, 9]:
    print(random.randint(0, k//2))
    print(random.randint(k//2, k))
\end{lstlisting}

\section{Evaluation on Individual Metrics}
\label{sec:appendix}

% 1          67.766509
% 2          60.379483
% 3          59.711503
% 4          50.935238
% 5          50.408485
% 6          50.235714
% 7          54.487778
% overall    60.800131

% 64.190828	57.000862	53.925948	52.230000	49.852727	57.217143	55.781111	57.561362

% 68.606154	60.259483	60.347712	56.260000	53.166061	54.101429	61.066667	62.018955

% {'category': 2,
%  'cite': 'purdue_writing_lab_2021',
%  'concise': ["Working as a photo technician's apprentice was an educational experience."],
%  'link': 'https://owl.purdue.edu/owl/general_writing/academic_writing/reverse_paramedic_method.html',
%  'wordy': 'Working as a pupil under someone who develops photos was an experience that really helped me learn a lot.'}
% 84.  , 45.42

\begin{table*}[t!]
\begin{subtable}{.55\linewidth}\centering{
\begin{tabular}{llllllll}
\hline
          &\textbf{BL}   & \textbf{M}   & \textbf{R}   & \textbf{S}   & \textbf{P}  & \textbf{BS}   & \textbf{T}  \\\hline
ParaNMT   & .40          & .78          & .56          & .49          & .55          & .96          & .53          \\
MSR       & .55          & .86          & .69          & .62          & .73          & \textbf{.97} & .39          \\
News      & .36          & .67          & .56          & .52          & .62          & .95          & .51          \\
Gigaword  & .04          & .27          & .19          & .25          & .16          & .89          & .92          \\
WikiSmall & .47          & \textbf{.91} & .65          & .59          & .63          & .95          & .82          \\
Approach 1   & .48          & .90          & .65          & .59          & .66          & .94          & .82          \\
%Input     & .48          & \textbf{.93} & .66          & \textbf{.68} & .68          & .96          & .56          \\
Approach 2  & \textbf{.57} & .87          & \textbf{.71} & \textbf{.66}  & \textbf{.74} & \textbf{.97} & \textbf{.37} \\ \hline
\end{tabular}
}
\caption{Category \RomanNumeralCaps{1}, 169 / 536, Delete}
\end{subtable}
\begin{subtable}{.42\linewidth}\centering{
\begin{tabular}{lllllll}
\hline
\textbf{BL}  & \textbf{M}   & \textbf{R}   & \textbf{S}   & \textbf{P}  & \textbf{BS}   & \textbf{T}  \\
\hline
\textbf{.38} & .75          & \textbf{.54} & \textbf{.54} & \textbf{.51} & \textbf{.97} & \textbf{.50} \\
.33          & .73          & .51          & .45          & .47          & .96          & .56          \\
.17          & .56          & .39          & .35          & .37          & .94          & .64          \\
.01          & .26          & .17          & .28          & .15          & .89          & .92          \\
.33          & .80          & .51          & .48          & .44          & .95          & .92          \\
.36          & \textbf{.80} & \textbf{.54} & .51          & .46          & .95          & .91          \\
% .36          & \textbf{.82} & .53          & .52          & .47          & .96          & .65          \\
.36          & .76          & .53          & .50          & .50          & .96          & .53          \\ \hline
\end{tabular}
}
\caption{Category \RomanNumeralCaps{2}, 116 / 536, Replace}
\end{subtable}

\vspace{3mm}
\begin{subtable}{.55\linewidth}\centering{
\begin{tabular}{llllllll}
\hline
ParaNMT   & .16          & .60          & .33          & .44          & .28          & \textbf{.94} & .94          \\
MSR       & .15          & .56          & .31          & .38          & .28          & .93          & .92          \\
News      & .10          & .44          & .26          & .37          & .25          & .92          & .85          \\
Gigaword  & .03          & .20          & .12          & .38          & .09          & .88          & 1.00          \\
WikiSmall & \textbf{.19} & \textbf{.66 }& .35          & \textbf{.45} & .29          & .93          & 1.26          \\
Approach 1   & .17          & .65          & \textbf{.36} & .43          & .29          & .93          & 1.27          \\
%Input     & \textbf{.19} & \textbf{.67} & \textbf{.37} & \textbf{.47} & \textbf{.31} & \textbf{.94} & 1.09          \\
Approach 2  & .18          & .60          & .35          & .42          & \textbf{.30 }  & \textbf{.94} & \textbf{.91} \\ \hline
\end{tabular}
}
\caption{Category \RomanNumeralCaps{3}, 153 / 536, Rewrite}
\end{subtable}
\begin{subtable}{.42\linewidth}\centering{
\begin{tabular}{lllllll}
\hline
.27          & .68          & .40          & \textbf{.48} & .41          & .94          & 1.22          \\
.26          & .63          & .39          & .44          & .39          & .94          & 1.09          \\
.12          & .41          & .29          & .37          & .31          & .92          & \textbf{.77} \\
.03          & .24          & .16          & .35          & .16          & .89          & .86          \\
.24          & .71          & .39          & .46          & .37          & .93          & 1.69          \\
.23          & \textbf{.72} & .39          & .44          & .39          & .92          & 1.70          \\
% .24          & .72          & .39          & .48          & .39          & .94          & 1.58          \\
\textbf{.32} & .67          & \textbf{.44} & .47          & \textbf{.43} & \textbf{.95} & .94          \\ \hline
\end{tabular}
}
\caption{Category \RomanNumeralCaps{4}, 42 / 536, Delete + Replace}
\end{subtable}

\vspace{3mm}
\begin{subtable}{.55\linewidth}\centering{
\begin{tabular}{llllllll}
\hline
ParaNMT   & .16          & .55          & .27          & \textbf{.44} & .26          & \textbf{.94} & 1.18          \\
MSR       & .14          & .49          & .28          & .37          & .24          & .93          & 1.05          \\
News      & .08          & .34          & .21          & .35          & .22          & .92          & \textbf{.85} \\
Gigaword  & .01          & .15          & .09          & .34          & .06          & .87          & .90          \\
WikiSmall & .17          & .58          & .30          & .40          & .27          & .93          & 1.40          \\
Approach 1   & \textbf{.20}   & \textbf{.59} & \textbf{.31} & .43          & \textbf{.29} & .93     & 1.37          \\
%Input     & \textbf{.19} & \textbf{.59} & \textbf{.31} & \textbf{.45} & \textbf{.30} & \textbf{.94} & 1.31          \\
Approach 2  & .14          & .49          & .26          & .35          & .25          & \textbf{.93}  & 1.04          \\ \hline
\end{tabular}
}
\caption{Category \RomanNumeralCaps{5}, 33 / 536, Replace + Rewrite}
\end{subtable}
\begin{subtable}{.42\linewidth}\centering{
\begin{tabular}{lllllll}
\hline
.21          & .56          & .30          & \textbf{.43} & .27          & \textbf{.93} & 1.35          \\
.19          & .53          & .30          & .40          & .30          & \textbf{.93} & 1.23          \\
.12          & .39          & .26          & .37          & .25          & .92          & \textbf{.81} \\
.04          & .20          & .12          & .38          & .10          & .87          & .87          \\
.17          & .62          & \textbf{.31} & .43          & \textbf{.30} & .92          & 1.92          \\
.16          & \textbf{.62} & \textbf{.31} & .41          & .29          & .92          & 1.96          \\
% .18          & \textbf{.63} & \textbf{.31} & \textbf{.45} & \textbf{.32} & \textbf{.93} & 1.74          \\
\textbf{.22} & .55          & \textbf{.31} & .41          & \textbf{.30} & \textbf{.93} & 1.23          \\ \hline
\end{tabular}
}
\caption{Category \RomanNumeralCaps{6}, 14 / 536, Delete + Rewrite}
\end{subtable}

\vspace{3mm}
\begin{subtable}{.55\linewidth}\centering{
\begin{tabular}{llllllll}
\hline
ParaNMT   & .06          & .50          & .17          & .40          & .20          & .92          & 1.61          \\
MSR       & .06          & .49          & .17          & .39          & .17          & .92          & 1.34          \\
News      & .03          & .28          & .15          & .38          & .21          & .91          & \textbf{.79} \\
Gigaword  & .00          & .11          & .04          & .36          & .02          & .86          & .95          \\
WikiSmall & \textbf{.08} & .54          & .20          & .38          & .18          & .91          & 2.02          \\
Approach 1   & .04          & .53          & .18          & .39          & .17          & .91          & 1.96          \\
%Input     & .06          & .54          & .18          & .41          & .17          & .92          & 1.84          \\
Approach 2  & \textbf{.08} & \textbf{.55} & \textbf{.23} & \textbf{.43} & \textbf{.24} & \textbf{.93} & 1.18          \\ \hline
\end{tabular}
}
\caption{Category \RomanNumeralCaps{7}, 9 / 536, Delete + Replace + Rewrite}
\end{subtable}
\begin{subtable}{.42\linewidth}\centering{
\begin{tabular}{lllllll}
\hline
.29          & .69          & .44          & .48          & .42          & \textbf{.95} & .77          \\
.32          & .69          & .48          & .48          & .47          & \textbf{.95} & .71          \\
.20          & .53          & .38          & .41          & .40          & .93          & .69          \\
.03          & .23          & .15          & .31          & .13          & .88          & .94          \\
.31          & .76          & .48          & .49          & .43          & .94          & 1.12          \\
.31          & \textbf{.76 } & .49          & .50          & .45          & .94          & 1.12          \\
% .32          & \textbf{.78} & .49          & \textbf{.54} & .47          & \textbf{.95} & .91          \\
\textbf{.35} & .72          & \textbf{.50} & \textbf{.51} & \textbf{.49} & \textbf{.95} & \textbf{.68} \\ \hline
\end{tabular}
}
\caption{Overall}
\end{subtable}
\caption{BLEU (\textbf{BL}), METEOR (\textbf{M}) , ROUGE-2-F1 (\textbf{R}), SARI (\textbf{S}), Parsed relation F1 (\textbf{P}), BERTScore-F1 (\textbf{BS}), \textit{and} translation edit rate (\textbf{T}) of pre-Approach 2s and Approach 2 method. Numbers are shown in categories. Smaller edit distance is more favorable. The most favorable score(s) in each column is bold. In category \RomanNumeralCaps{5}, the model trained on Approach 1 has the highest scores on three metrics, slightly outperforming other pre-Approach 2s. In category \RomanNumeralCaps{3},\RomanNumeralCaps{4}, \RomanNumeralCaps{6}, \RomanNumeralCaps{7}, no particular pre-Approach 2 scores well on all metrics.}\label{tab:results}
\end{table*}

\begin{table*}[t!]
\begin{subtable}{.43\linewidth}\centering{
\begin{tabular}{llll}
\hline
          &\textbf{W}   & \textbf{R1}   & \textbf{RL}\\
\hline
ParaNMT   & .65          & .75          & .71          \\
MSR       & .43          & .85          & .8           \\
News      & .54          & .73          & .7           \\
Gigaword  & .97          & .39          & .37          \\
WikiSmall & .89          & .82          & .77          \\
Approach 1   & .96          & .82          & .77          \\
% Input     & .63          & .83          & .78          \\
Approach 2  & \textbf{.42} & \textbf{.86} & \textbf{.81} \\ \hline
\end{tabular}
}
\caption{Category \RomanNumeralCaps{1}, 169 / 536, Delete}
\end{subtable}
\begin{subtable}{.41\linewidth}\centering{
\begin{tabular}{lll}
\hline
\textbf{W}   & \textbf{R1}   & \textbf{RL}  \\
\hline
.61          & \textbf{.73} & \textbf{.72} \\
.58          & .71          & .7           \\
.66          & .6           & .59          \\
.98          & .37          & .35          \\
.93          & .71          & .7           \\
1.03          & \textbf{.73} & \textbf{.72} \\
% .65          & .72          & .71          \\
\textbf{.54} & \textbf{.73} & \textbf{.72} \\ \hline
\end{tabular}
}
\caption{Category \RomanNumeralCaps{2}, 116 / 536, Replace}
\end{subtable}

\vspace{3mm}
\begin{subtable}{.43\linewidth}\centering{
\begin{tabular}{llll}
\hline
ParaNMT   & 1.04          & .61          & .50          \\
MSR       & .99          & .60          & .48          \\
News      & .88          & .52          & .43          \\
Gigaword  & 1.03          & .32          & .29          \\
WikiSmall & 1.35          & .63          & .50          \\
Approach 1   & 1.40          & .63          & .50          \\
% Input     & 1.18          & \textbf{.64} & \textbf{.51} \\
Approach 2  & \textbf{.97} & .63          & \textbf{.51} \\ \hline
\end{tabular}
}
\caption{Category \RomanNumeralCaps{3}, 153 / 536, Rewrite}
\end{subtable}
\begin{subtable}{.41\linewidth}\centering{
\begin{tabular}{lll}
\hline
1.28          & .59          & .58          \\
1.14          & .59          & .57          \\
\textbf{.81} & .48          & .46          \\
.90          & .37          & .32          \\
1.74          & .58          & .56          \\
1.79          & .59          & .56          \\
% 1.62          & .57          & .56          \\
.97          & \textbf{.63} & \textbf{.62} \\ \hline
\end{tabular}
}
\caption{Category \RomanNumeralCaps{4}, 42 / 536, Delete + Replace}
\end{subtable}

\vspace{3mm}
\begin{subtable}{.43\linewidth}\centering{
\begin{tabular}{llll}
\hline
ParaNMT   & 1.36          & .52          & .45          \\
MSR       & 1.17          & .51          & .43          \\
News      & \textbf{.91} & .42          & .36          \\
Gigaword  & .96          & .28          & .24          \\
WikiSmall & 1.58          & .53          & .44          \\
Approach 1   & 1.59          & \textbf{.55} & \textbf{.46} \\
% Input     & 1.48          & .54          & .45          \\
Approach 2  & 1.15          & .52          & .43          \\ \hline
\end{tabular}
}
\caption{Category \RomanNumeralCaps{5}, 33 / 536, Replace + Rewrite}
\end{subtable}
\begin{subtable}{.41\linewidth}\centering{
\begin{tabular}{lll}
\hline
1.54          & .49          & .41          \\
1.37          & .51          & .41          \\
\textbf{.87} & .45          & .38          \\
.89          & .30          & .30          \\
2.04          & .51          & .43          \\
2.10          & .51          & .43          \\
% 1.89          & .51          & .44          \\
1.39          & \textbf{.54} & \textbf{.45} \\ \hline
\end{tabular}
}
\caption{Category \RomanNumeralCaps{6}, 14 / 536, Delete + Rewrite}
\end{subtable}

\vspace{3mm}
\begin{subtable}{.43\linewidth}\centering{
\begin{tabular}{llll}
\hline
ParaNMT   & 1.64          & .44          & .35          \\
MSR       & 1.38          & .47          & .35          \\
News      & \textbf{.81} & .38          & .34          \\
Gigaword  & .97          & .25          & .20          \\
WikiSmall & 2.04          & .44          & .33          \\
Approach 1   & 1.98          & .45          & .34          \\
% Input     & 1.88          & .44          & .33          \\
Approach 2  & 1.24          & \textbf{.53} & \textbf{.43} \\ \hline
\end{tabular}
}
\caption{Category \RomanNumeralCaps{7}, 9 / 536, Delete + Replace + Rewrite}
\end{subtable}
\begin{subtable}{.41\linewidth}\centering{
\begin{tabular}{lll}
\hline
.88          & .66          & .61          \\
.76          & .69          & .63          \\
\textbf{.72} & .59          & .55          \\
.98          & .35          & .33          \\
1.19          & .69          & .63          \\
1.25          & .70          & .63          \\
% .98          & .70          & .63          \\
.73          & \textbf{.71} & \textbf{.65} \\ \hline
\end{tabular}
}
\caption{Overall}
\end{subtable}
\caption{Other metrics include word error rate (\textbf{W}), ROUGE-1-F1 (\textbf{R1}), \textit{and} ROUGE-L-F1 (\textbf{RL}) . Numbers are shown in categories. Smaller edit distance is more favorable. The most favorable score(s) in each column is bold.}\label{tab:results_appendix}
\end{table*}

For each sample in the benchmark corpus, we compute individual metric score for its best-revised sentence and average the corpus ranking of its individual metric scores to obtain the final ranking for that sample. Table~\ref{tab:good} lists the well-treated samples (at the third percentile) in each category. Table~\ref{tab:bad} lists the cases that were not well resolved (at the 97th percentile). 

\begin{table*}[!t]
\centering
\begin{tabularx}{\textwidth}{lXX}
\hline
\textbf{Category} & \textbf{Reference(s)} & \textbf{Prediction} \\\hline
\RomanNumeralCaps{1} & \remove{The reason} she moved \remove{is} because she was offered a better position. & \remove{The reason} she moved \remove{is} because she was offered a better position. \\
\rowcolor{mygray}
\RomanNumeralCaps{2} & We outlined the letters for the banner \replace{in a careful way}{carefully}. & We outlined the letters for the banner \replace{in a careful way}{carefully}. \\
\RomanNumeralCaps{3} & \replace{The current focus of the medical profession is}{The medical profession currently focuses on} disease prevention. & \replace{The current focus of the medical profession is}{The medical profession focuses on} disease prevention.\\
\rowcolor{mygray}
\RomanNumeralCaps{4} & She fell \remove{down} \replace{due to the fact that}{because}  she hurried. & She fell \remove{down} \replace{due to the fact that}{because}  she hurried. \\
\RomanNumeralCaps{5} & We didn't renew the contract with our supplier \replace{on account of the fact that}{because} \replace{it was necessary for us to}{we had to} find a cheaper \replace{supplier}{one}. & We didn't renew the contract \remove{with our supplier} \replace{on account of the fact that}{because} it was necessary for us to find a cheaper supplier. \\
\rowcolor{mygray}
\RomanNumeralCaps{6} & Research is increasing in \remove{the field of} nutrition and food science. \emph{OR} Research \add{within nutrition and food science} is increasing \remove{the field of nutrition and food science}. & Research is increasing in \remove{the field of} nutrition and food science. \\
\RomanNumeralCaps{7} & Joe found \remove{that} \remove{the fictional novel by} Alcott\add{'s book}, Little Women, \replace{conveyed a sense of emotion and feeling}{both delightful and tragic}. & Joe found that \remove{the fictional novel by} Alcott\add{'s novel}, Little Women, conveyed \remove{a sense of} emotion and feeling. \\
\hline
\end{tabularx}
\caption{\label{tab:good}
Well-revised samples in each category (about the third percentile). Shorter sentences that require simpler actions are perfectly revised. When sentence length increases, not all wordiness can be resolved. The baseline method tends to delete and replace, even if sentences need to be rewritten.
}
\end{table*}

\begin{table*}[t!]
\centering
\begin{tabularx}{\textwidth}{lXX}
\hline
\textbf{Category} & \textbf{Reference(s)} & \textbf{Prediction} \\\hline
\RomanNumeralCaps{1} & \remove{I would call your attention to the fact that} our President, \remove{who was} formerly the Governor of Arkansas, is \remove{basically} a Southerner. & I would call your attention to the fact that our President \remove{, who was formerly the Governor of Arkansas,} is a \remove{basically} Southerner. \\
\rowcolor{mygray}
\RomanNumeralCaps{2} & We have \replace{conducted an investigation}{investigated} and \replace{arrived at the conclusion}{concluded}. & We \remove{have} conducted an investigation and arrived at the conclusion. \\
\RomanNumeralCaps{3} & \replace{As far as my professor is concerned, the problem of wordiness is the thing she'd really like to see us involved with actually eliminating.}{My professor wants us to focus on eliminating wordiness.} & \remove{As far as my professor is concerned,} the problem of wordiness is the thing she'd \remove{really} like to see us involved with \remove{actually} eliminating.\\
\rowcolor{mygray}
\RomanNumeralCaps{4} & \remove{The existence of} computers \remove{and computer technology} \replace{has}{have} {improved} \replace{commercial enterprise}{business} \remove{and} \replace{information exchange}{communication}. & The existence of computers and \remove{computer} technology has \remove{greatly} influenced commercial enterprise and information exchange. \\
\RomanNumeralCaps{5} & \replace{It has come to my attention that}{I have noticed} \replace{there is a vast proliferation of undesirable vegetation}{many weeds growing} \replace{surrounding the periphery of}{around} \replace{this facility}{the building} & It has come to my attention that there is a vast proliferation of undesirable vegetation \remove{surrounding the periphery of this facility}. \\
\rowcolor{mygray}
\RomanNumeralCaps{6} & \replace{Your statement in your letter that has the contention that the information submitted and sent to you by us contained certain inaccuracies and errors has prompted us to embark on a careful and thorough reevaluation of the information submitted, with the result that the original informative data has been determined to be accurate and correct in all instances and aspects of the information.}{As you suggested, we have checked our information and confirmed its accuracy.} & Your statement in your letter that \remove{has the contention that} the information submitted and sent to you by us contained \replace{certain}{some} inaccuracies and errors has prompted us to embark on a \remove{careful and} thorough reevaluation of the information \remove{submitted}, with the result that the original informative data has been determined to be accurate and correct \remove{in all instances and aspects of the information}.
 \\
\RomanNumeralCaps{7} & \replace{In the event that}{If} \add{you get} \replace{some}{any} information \replace{concerning}{about} Mr. Smith \remove{should be brought to your attention}, \replace{it should be forwarded via mail or courier or telephone to us}{please contact us} \replace{in view of the possibility that}{in case} \replace{the information may reveal any attempt on the part of Mr. Smith to depart from the United States}{he tries to leave the country}. &	\replace{In the event that}{If} \replace{some}{any} information concerning Mr. Smith should be brought to your attention, it should be forwarded via mail or courier or telephone to us in view of the possibility that the information may reveal any attempt \remove{on the part of Mr. Smith} to depart from the United States.
 \\
\hline
\end{tabularx}
\caption{\label{tab:bad}
Badly-revised samples in each category (about the 97th percentile). These sentences are longer than sentences in Table~\ref{tab:good}. Informative part may be trimmed. Replacing nominalizations with verbs is hard. For severely wordy sentences (category \RomanNumeralCaps{6}, \RomanNumeralCaps{7}), the model fails to rewrite, and resorts to deletion. A lot of improvement is needed.
}
\end{table*}

\begin{figure}[t]
    \centering
    \includegraphics[width=0.5\textwidth]{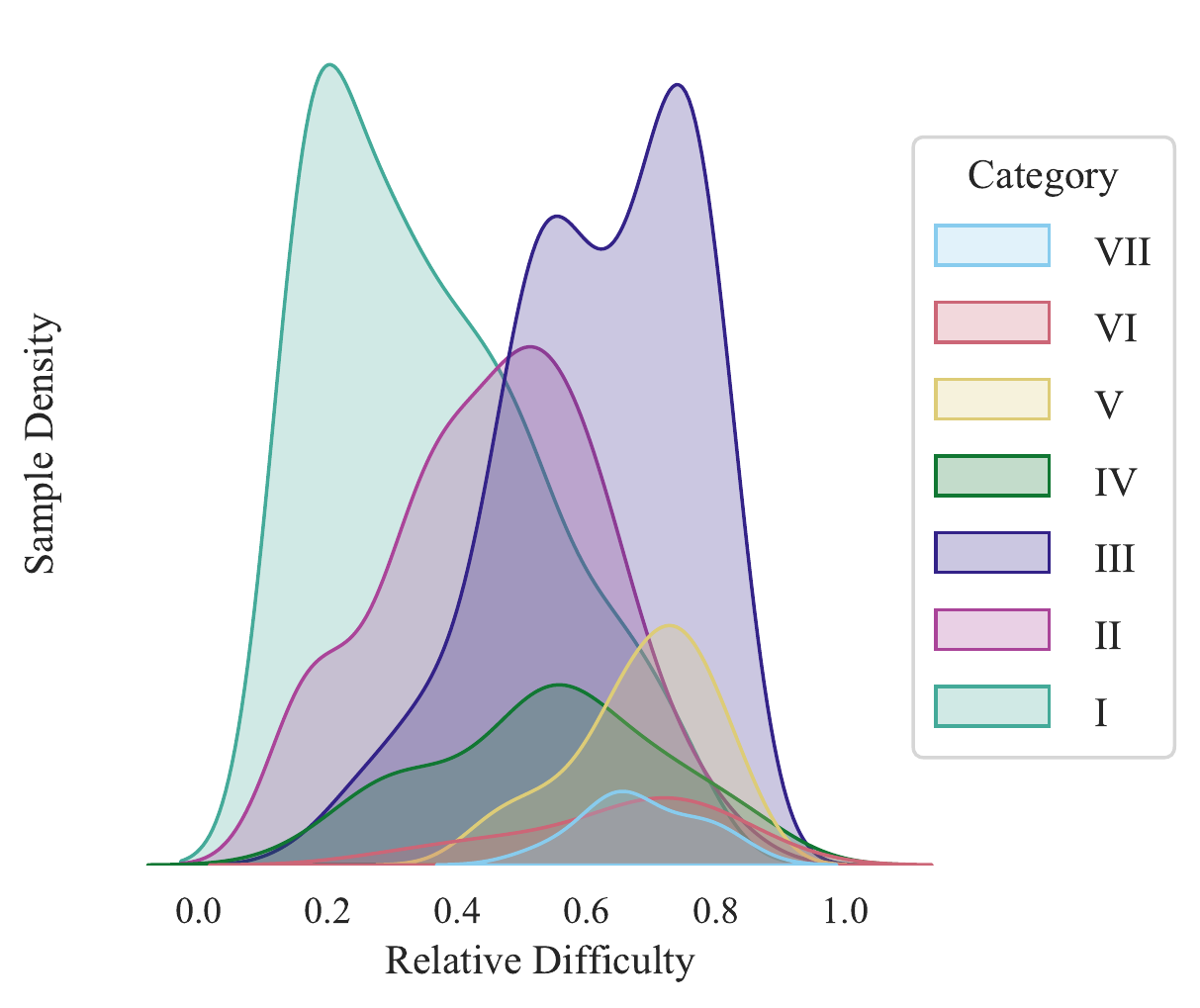}
    \caption{Difficulty faced by the proposed approachs when dealing with sentences from different categories. This difficulty is relative to other samples in the corpus of 536 sentences. Deletion (category \RomanNumeralCaps{1}) is the least challenging. The most challenging samples are most likely from category \RomanNumeralCaps{3}. Handling sentences requiring more than one revising strategies (category \RomanNumeralCaps{4}-\RomanNumeralCaps{7}) is usually more challenging.}
    \label{fig:difficulty}
\end{figure}

Figure~\ref{fig:difficulty} shows the difficulty of revising sentences for each category. The data in Figure~\ref{fig:difficulty}, while demonstrating strengths and weaknesses of the proposed approach, can also serve as an approximation of the difficulty of the corpus itself. The proposed approach is better at deleting and replacing than rewriting due to heavy reliance on transfer learning.

\end{document}